\documentclass{article}
\usepackage{graphicx}
\usepackage{hyperref}
\usepackage{authblk}

\begin{document}

\title{Solving classification problems using Traceless Genetic Programming}

\author{Mihai Oltean\\
Department of Computer Science,\\
Faculty of Mathematics and Computer Science,\\
Babes-Bolyai University, Kogalniceanu 1,\\
Cluj-Napoca, 3400, Romania.\\
mihai.oltean@gmail.com
}

\date{}
\maketitle

\begin{abstract}

Traceless Genetic Programming (TGP) is a new Genetic Programming (GP) that may be used for solving difficult real-world problems. The main difference between TGP and other GP techniques is that TGP does not explicitly store the evolved computer programs. In this paper TGP is used for solving real-world classification problems taken from PROBEN1. Numerical experiments show that TGP performs similar and sometimes even better than other GP techniques for the considered test problems.

\end{abstract}

\section{Introduction}

Classification is a task of assigning inputs to a number of discrete 
categories or classes \cite{haykin1}. Examples include classifying a handwritten 
letter as one from A-Z, classifying a speech pattern to the corresponding 
word, etc.

Machine learning techniques have been extensively used for solving 
classification problems. In particular Artificial Neural Networks (ANNs) \cite{haykin1} 
have been originally designed for classifying a set of points in two 
distinct classes. 

Evolutionary Computation \cite{goldberg1} and recently Genetic Programming (GP) technique \cite{koza1} have been used for 
classification purposes. GP has been developed as a technique for evolving 
computer programs. Originally GP chromosomes were represented as trees and 
evolved using specific genetic operators \cite{koza1}. 

Some of the GP variants have been used for classification purposes. For 
instance, Linear Genetic Programming \cite{brameier1} has been used for solving several classification 
problems in PROBEN1. The conclusion was that LGP is able to solve the 
classification problems with the same error rate as a neural network.

In this paper Traceless Genetic Programming \cite{oltean1} is used for solving several real-world classification problems taken from PROBEN1.

Traceless Genetic Programming (TGP) \footnote {The source code for TGP is available at \url{https://github.com/mihaioltean/traceless-genetic-programming}} is a GP \cite{koza1} variant as it evolves a population of computer programs. The main difference between the TGP and GP is that TGP does not explicitly store the evolved computer programs. TGP is useful when the trace (the way in which the results are obtained) between the input and output is not important. In this way the space used by traditional techniques for storing the entire computer programs (or mathematical expressions in the simple case of symbolic regression) is saved.

We choose to apply the TGP technique to the classification problems due to their high practical interest and easy description.

The paper is organized as follows: Two machine learning techniques used for solving classification problems are presented in section \ref{ml}. In section \ref{tgp} the Traceless Genetic Programming technique is briefly described. Data sets used for assessing the performance of the considered algorithms are presented in section \ref{data_sets}. In section \ref{exp} several numerical experiments for solving classification problems are performed.

\section{Machine learning techniques used for classification task}\label{ml}

Two of the machines learning techniques used for solving classification 
problems are described in this section.

\subsection{Artificial Neural Networks }

Artificial Neural Networks (ANNs) \cite{haykin1} are motivated by biological 
neural networks. ANNs have been extensively used for classification and 
clustering purposes. In their early stages, neural networks have been 
primarily used for classification tasks. The most popular architectures 
consist of multi layer perceptrons (MLP) and the most efficient training 
algorithm is backpropagation \cite{prechelt1}.

ANNs have been successfully applied for solving real-world problems. For 
instance, the PROBEN1 \cite{prechelt1} data set contains several difficult real-world 
problems accompanied by the results obtained by running different ANNs 
architectures on the considered test problems. The method applied in PROBEN 
for training was RPROP (a fast backpropagation algorithm - similar to 
Quickprop). The parameters used were not determined by a trial-and-error 
search. Instead they are just educated guesses. More information about the 
ANNs parameters used in PROBEN1 can be found in \cite{prechelt1}.

\subsection{Linear Genetic Programming}

{\it Linear Genetic Programming} (LGP) \cite{brameier1} uses a specific linear representation of computer programs. 

Programs of an imperative language (like \textbf{\textit{C}}) are evolved 
instead of the tree-based GP expressions of a functional programming 
language (like \textbf{\textit{LISP}})

An LGP individual is represented by a variable-length sequence of simple 
\textbf{\textit{C}} language instructions. Instructions operate on one or 
two indexed variables (registers) $r$ or on constants $c$ from predefined sets. 
The result is assigned to a destination register.

An example of LGP program is the following:\\

\textsf{\textbf{void}}\textsf{ 
LGP{\_}Program(}\textsf{\textbf{double}}\textsf{ v[8])}

\textsf{{\{}}

\hspace{1cm}\textsf{\ldots }

\hspace{1cm}\textsf{v[0] = v[5] + 73;}

\hspace{1cm}\textsf{if (v[5] $>$ 21)}

\hspace{1cm}\textsf{v[4] = v[2] *v[1];}

\hspace{1cm}\textsf{v[2] = v[5] + v[4];}

\hspace{1cm}\textsf{v[6] = v[1] * 25;}

\hspace{1cm}\textsf{v[6] = v[4] - 4;}

\hspace{1cm}\textsf{v[1] = sin(v[6]);}

\hspace{1cm}\textsf{if (v[0] $>$ v[1])}

\hspace{1cm}\textsf{v[3] = v[5] * v[5];}

\textsf{\}
}\\

A linear genetic program can be turned into a functional representation by 
successive replacements of variables starting with the last effective 
instruction. Variation operators are crossover and mutation. By crossover 
continuous sequences of instructions are selected and exchanged between 
parents. Two types of mutations are used: micro mutation and macro mutation. 
By micro mutation an operand or operator of an instruction is changed. Macro 
mutation inserts or deletes a random instruction.

Crossover and mutations change the number of instructions in chromosome. 

The population is divided into demes which are arranged in a ring. Migration 
is allowed only between a deme and the next one (in the clockwise direction) 
with a fixed probability of migration. Migrated individuals replace the 
worst individuals in the new deme. The deme model is used in order to 
preserve population diversity (see \cite{brameier1} for more details).

\section{Traceless Genetic Programming}\label{tgp}

In this section the TGP technique \cite{oltean1} is briefly described. 

\subsection{Prerequisite}

The quality of a GP individual is usually computed using a set of fitness cases \cite{koza1}. For instance, the aim of symbolic regression is to find a mathematical expression that satisfies a set of $m$ fitness cases. 

We consider a problem with $n$ inputs: $x_{1}$, $x_{2}$, \ldots $x_{n}$ and one output $f$. The inputs are also called terminals \cite{koza1}. The function symbols that we use for constructing a mathematical expression are $F=\{+,-,*,/, sin\}$.

Each fitness case is given as a ($n+1$) dimensional array of real values:\\

\[
v_1^k ,v_2^k ,v_3^k ,...,v_n^k ,f_k 
\]

\noindent
where $v_j^k $ is the value of the $j^{th}$ attribute (which is $x_{j})$ in 
the $k^{th}$ fitness case and $f_{k}$ is the output for the $k^{th}$ fitness 
case.

Usually more fitness cases are given (denoted by $m$) and the task is to find the expression that satisfies best all these fitness cases. This is usually done by minimizing the quantity:

\[
Q=\sum\limits_{k = 1}^m {\left| {f_k - o_k } \right|} ,
\]

\noindent
where $f_{k}$ is the target value for the $k^{th}$ fitness case and $o_{k}$ is 
the actual (obtained) value for the $k^{th}$ fitness case.

\subsection{Individual representation}

Each TGP individual represents a mathematical expression evolved so far, but 
the TGP individual does not explicitly store this expression. Each TGP 
individual stores only the obtained value, so far, for each fitness case. Thus 
a TGP individual is:\\

($o_{1}$, $o_{2}$, $o_{3}$, \ldots , $o_{m})^{T}$,\\

\noindent
where $o_{k}$ is the current value for the $k^{th}$ fitness case. Each 
position in this array (a value $o_{k})$ is a gene. As we said it before 
behind these values is a mathematical expression whose evaluation has 
generated these values. However, we do not store this expression. We store 
only the values $o_{k}$.\\

\textbf{Remark}

The structure of an TGP individual can be easily enhanced for storing the evolved computer program (mathematical expression). Storing the evolved expression can provide a more easy way to analyze the results of the numerical experiments. However, in this paper, we do not store the trees associated with the TGP individuals.

\subsection{Initial population}

The initial population contains individuals whose values have been generated 
by simple expressions (made up a single terminal). For instance, if an 
individual in the initial population represent the expression:\\

$E=x_{1}$,\\

\noindent
then the corresponding TGP individual is represented as:\\

\[
C = (v_1^1 ,v_1^2 ,v_1^3 ,...,v_1^m )^T
\]

\noindent
where $v_j^k $ has been previously explained.

The quality of this individual is computed using the equation previously 
described:

\[
Q = \sum\limits_{i = 1}^m {\left| {v_1^k - f_k } \right|} .
\]

\subsection{Genetic Operators}

In this section the genetic operators used in conjunction with TGP are 
described. TGP uses two genetic operators: crossover and insertion. These operators are specially designed for the proposed TGP technique.

\subsubsection{Crossover}

The crossover is the only variation operator that creates new individuals. For 
crossover several individuals (the parents) and a function symbol are 
selected. The offspring is obtained by applying the selected 
operator for each of the genes of the parents.

The number of parents selected for crossover depends on the number of 
arguments required by the selected function symbol. Two parents have to be selected for crossover if the function symbol 
is a binary operator. A single parent needs to be selected if the function symbol is a unary operator.\\

\textbf{Example 1}\\

Let us suppose that the operator + is selected. In this case two parents: \\

$C_{1}$ = ($p_{1}$, $p_{2}$, \ldots , $p_{m})^{T}$ and

$C_{2}$ = ($q_{1}$, $q_{2}$, \ldots , $q_{m})^{T}$\\

\noindent
are selected and the offspring $O$ is obtained as follows:\\

$O$ = ($p_{1}+q_{1}$, $p_{2}+q_{2}$,\ldots , $p_{m}+q_{m})^{T}$.\\

\textbf{Example 2}\\

Let us suppose that the operator \textit{sin} is selected. In this case one parent:\\

$C_{1}$ = ($p_{1}$, $p_{2}$, \ldots , $p_{m})^{T}$ \\

\noindent
is selected and the offspring $O$ is obtained as follows:\\

$O$ = (\textit{sin}($p_{1})$, \textit{sin}($p_{2})$,\ldots , \textit{sin}($p_{m}))^{T}$.\\

\subsubsection{Insertion}

This operator inserts a simple expression (made up a single terminal) in the population. This operator is useful when the population contains individuals representing very complex expressions that cannot improve the search. By inserting simple expressions we give a chance to the evolutionary process to choose another direction for evolution.

\subsection{TGP Algorithm}

Due to the special representation and due to the special genetic 
operators, TGP uses a special generational algorithm which is given below:

The TGP algorithm starts by creating a random population of individuals. The evolutionary process is run for a fixed number of generation. At each generation the following steps are repeated until the new population is filled: With a probability $p_{insert}$ generate an offspring made of a single terminal (see the Insertion operator). With a probability 1-$p_{insert}$ select two parents using a standard selection procedure. The parents are recombined in order to obtain an offspring. 
The offspring enters the population of the next generation. 

The standard TGP algorithm is outlined below:\\

\begin{center}
\textbf{Standard TGP Algorithm}
\end{center}

\textsf{S}$_{1}$\textsf{. Randomly create the initial population 
}\textsf{\textit{P}}\textsf{(0)}

\textsf{S}$_{2.}$\textsf{\textbf{ for}}\textsf{ 
}\textsf{\textit{t}}\textsf{ = 1 }\textsf{\textbf{to}}\textsf{ 
}\textsf{\textit{NumberOfGenerations}}\textsf{ }\textsf{\textbf{do}}

\textsf{S}$_{3. }$\hspace{0.5cm}\textsf{\textit{P}}\textsf{'(}\textsf{\textit{t}}\textsf{) = $\phi $}

\textsf{S}$_{4.}$\hspace{0.5cm}\textsf{Copy the best individual from 
}\textsf{\textit{P}}\textsf{(}\textsf{\textit{t}}\textsf{) to 
}\textsf{\textit{P}}\textsf{'(}\textsf{\textit{t}}\textsf{)}

\textsf{S}$_{5.}$\hspace{0.5cm}\textsf{\textbf{while 
}}\textsf{\textit{P}}\textsf{'(}\textsf{\textit{t}}\textsf{) is not 
filled}\textsf{\textbf{ do}}

\textsf{S}$_{6.}$\hspace{1cm}\textsf{With a fixed insertion probability 
}\textsf{\textit{p}}$_{insert}$\textsf{ do}

\textsf{S}$_{7.}$\hspace{1.5cm}\textsf{Create an offspring \textit{offspr} made up a single terminal}

\textsf{S}$_{8.}$\hspace{1cm}\textsf{With a crossover probability 1 -- 
}\textsf{\textit{p}}$_{insert}$\textsf{ do}

\textsf{S}$_{9.}$\hspace{1.5cm}\textsf{\textit{Select}}\textsf{ an function symbol}

\textsf{S}$_{10.}$\hspace{1.35cm}\textsf{\textit{Select}}\textsf{ a number of 
parents equal to the number of arguments of the selected operator.}

\textsf{S}$_{11.}$\hspace{1.35cm}\textsf{\textit{Crossover}}\textsf{ the selected 
parents. An offspring }\textsf{\textit{offspr}}\textsf{ is obtained }

\textsf{S}$_{12.}$\hspace{1cm}\textsf{Add the offspring \textit{offspr} to 
}\textsf{\textit{P}}\textsf{'(}\textsf{\textit{t}}\textsf{)}

\textsf{S}$_{13.}$\hspace{0.5cm}\textsf{\textbf{endwhile}}

\textsf{S}$_{14.}$\hspace{0.5cm}\textsf{\textit{P}}\textsf{(}\textsf{\textit{t}}\textsf{+1) = 
}\textsf{\textit{P}}\textsf{'(}\textsf{\textit{t}}\textsf{)}

\textsf{S}$_{15}$\textsf{. }\textsf{\textbf{endfor }}

\subsection{TGP for Classification Problems}

In this section we explain how the fitness of a TGP chromosome is computed.

Each class has associated a numerical value: the first class has the value 
0, the second class has the value 1 and the $m^{th}$ class has associated the 
numerical value $m - $1. Any other system of distinct numbers may be used. We 
denote by $d_{k}$ the number associated to the $k^{th}$ class. 

We have stored (in an TGP chromosome) the value $o_j$ of an expression $E$ 
for each row (example) $j$ in the training set. Then, each row in 
the training set will be classified to the nearest class (the class $k$ for 
which the difference $\vert d_k - o_k \vert$ is minimal). 

The fitness of a TGP chromosome is equal to the number of incorrectly 
classified examples in the training set. \\

\textbf{Remark}\\

Since the set of numbers associated with the problem classes was arbitrarily chosen it is expected that different systems of number to generate different solutions.

\subsection{Complexity of the TGP Decoding Process} \label{complexity}

A very important aspect of the GP techniques is the time complexity of the procedure used for computing the fitness of the newly created individuals.

The complexity of that procedure for the standard GP is: \\

$O(m*g)$,\\

where $m$ is the number of fitness cases and $g$ is average number of nodes in the GP tree \cite{koza1}.

By contrast, the TGP complexity is only \\

$O(m)$ \\

because the quality of a TGP individual can be computed by traversing it only once. The length of a TGP individual is $m$.

Due to this reason we may allow TGP programs to run $g$ times more generations in order to obtain the same complexity as the standard GP.

\section{Data sets}\label{data_sets}

Numerical experiments performed in this paper are based on several benchmark 
problems taken from PROBEN1 \cite{prechelt1}. These datasets were created based on the 
datasets from the UCI Machine Learning Repository \cite{uci1}.

Test problems are briefly described in what follows.

\textbf{Cancer} Diagnosis of breast cancer. Try to classify a tumor as either benignant or 
malignant based on cell descriptions gathered by microscopic examination.

\textbf{Diabetes} Diagnosis diabetes of Pima Indians based on personal data and the results of 
medical examinations try to decide whether a Pima Indian individual is 
diabetes positive or not.

\textbf{Heartc} Predicts heart disease. Decides whether at least one of four major vessels 
is reduced in diameter by more than 50{\%}. The binary decision is made 
based on personal data such as age sex smoking habits subjective patient 
pain descriptions and results of various medical examinations such as blood 
pressure and electro cardiogram results. This data set was originally created by Robert Detrano from V.A. Medical 
Center Long Beach and Cleveland Clinic Foundation.

\textbf{Horse} Predicts the fate of a horse that has colic. The results of a veterinary 
examination of a horse having colic are used to predict whether the horse 
will survive will die or should be euthanized.\\

The number of inputs, of classes and of available examples, for each test 
problem, are summarized in Table \ref{tab_data1}.

\begin{table}[htbp]
\caption{Summarized attributes of several classification problems from PROBEN1.}
\label{tab_data1}
\begin{center}
\begin{tabular}
{p{40pt}p{80pt}p{80pt}p{80pt}}
\hline
\textbf{Problem}& 
\textbf{Number of inputs}& 
\textbf{Number of classes }& 
\textbf{Number of examples} \\
\hline
\textbf{cancer}& 
9& 
2& 
699 \\
\textbf{diabetes}& 
8& 
2& 
768 \\
\textbf{heartc}& 
35& 
2& 
303 \\
\textbf{horse}& 
58& 
3& 
364 \\
\hline
\end{tabular}
\end{center}
\end{table}

\section{Numerical Experiments}\label{exp}

The results of several numerical experiments with ANNs, LGP and TGP are 
presented in this section.

Each data set is divided in three sub-sets (training set -50{\%}, validation 
set - 25 {\%}, and test set -- 25{\%}) \cite{prechelt1}.

The test set performance is computed for that chromosome which had minim 
validation error during the search process. This method, called \textit{early stopping}, is a good 
way to avoid overfitting \cite{prechelt1} of the population individuals to the particular 
training examples used. In that case the generalization performance will be 
reduced.

In \cite{brameier1} Linear GP was used to solve several classification problems from 
PROBEN1. Linear GP used a population of 5000 individuals, with maximum 256 instructions each, evolved for 250 generations. The function set was $F$ = {\{}+, -, *, /, \textit{sin}, \textit{exp}, if $>$, if $ \le ${\}} and the terminal set was $T$ = {\{}0,..,256{\}} $ \cup $ {\{}input variables{\}}.

Parameters of the TGP algorithm are given in Table \ref{tab_tgp1}.

\begin{table}[htbp]
\caption{General parameters of the TGP algorithm for solving the classification problems in PROBEN1.}
\label{tab_tgp1}
\begin{center}
\begin{tabular}
{p{100pt}p{100pt}}
\hline
\textbf{Parameter}& 
\textbf{Value} \\
\hline
Population size& 
500 \\
Number of generations& 
1000 \\
Insertion probability& 
0.2 \\
Selection& 
Binary Tournament \\
Function set& 
\{+, -, *, \/, sin, exp\} \\
Terminal set& 
input variables \\
Number of runs& 
30 \\
\hline
\end{tabular}
\end{center}
\end{table}

Results are given in Table \ref{tab_results}.

\begin{table}[htbp]
\caption{Classification error rates of TGP, LGP and ANN for some 
date sets from PROBEN1. LGP results are taken from \cite{brameier1}. ANNs results are 
taken from \cite{prechelt1}. Results are averaged over 30 runs.}
\label{tab_results}
\begin{tabular}
{p{30pt}p{25pt}p{25pt}p{25pt}p{25pt}p{25pt}p{25pt}p{25pt}p{25pt}}
\hline
\textbf{Problem}& 
\multicolumn{3}{p{80pt}}{\textbf{TGP--test set}} & 
\multicolumn{3}{p{80pt}}{\textbf{LGP--test set}} & 
\multicolumn{2}{p{60pt}}{\textbf{NN--test set}}  \\
\hline
 & 
\textbf{best}& 
\textbf{mean}& 
\textbf{stddev}& 
\textbf{best}& 
\textbf{mean}& 
\textbf{stddev}& 
\textbf{mean}& 
\textbf{stddev} \\
\hline
cancer1& 
0.57&
2.09&
1.07&
0.57& 
2.18& 
0.59& 
1.38& 
0.49 \\
cancer2& 
3.44&
5.45&
0.99&
4.02& 
5.72& 
0.66& 
4.77& 
0.94 \\
cancer3& 
2.87&
4.24&
0.98&
3.45& 
4.93& 
0.65& 
3.70& 
0.52 \\
diabetes1& 
23.95&
26.71&
1.80&
21.35& 
23.96& 
1.42& 
24.10& 
1.91 \\
diabetes2& 
25.00& 
28.12& 
1.71& 
25.00& 
27.85& 
1.49& 
26.42& 
2.26 \\
diabetes3& 
20.65& 
25.31& 
2.20& 
19.27& 
23.09& 
1.27& 
22.59& 
2.23 \\
heart1& 
17.33&
22.93&
3.72&
18.67& 
21.12& 
2.02& 
20.82& 
1.47 \\
heart2& 
2.66&
9.60&
2.99&
1.33& 
7.31& 
3.31& 
5.13& 
1.63 \\
heart3& 
13.33&
17.91&
2.68&
10.67& 
13.98& 
2.03& 
15.40& 
3.20 \\
horse1& 
26.37&
32.89&
3.06&
23.08& 
30.55& 
2.24& 
29.19& 
2.62 \\
horse2& 
27.47&
35.97&
3.86&
31.87& 
36.12& 
1.95& 
35.86& 
2.46 \\
horse3& 
30.76&
37.47&
2.93&
31.87&
35.44&
1.77& 
34.16& 
2.32 \\
\hline
\end{tabular}
\end{table}

Table \ref{tab_results} shows that TGP is able to obtain similar performances as 
those obtained by LGP even if the population size and the chromosome length 
used by TGP are smaller than those used by LGP. Better results can be obtained if an larger population and an increased number of generation are used. Note that one TGP generation is not equivalent with one LGP generation (see section \ref{complexity}).

\section{Conclusion and Further Work}

In this paper Traceless Genetic Programming has been applied for solving real-world classification problems. TGP has been compared to Linear GP and Neural Networks. Numerical experiments have shown that TGP was able to solve the considered problems with a similar error rate as NNs and LGP.
 
Further effort will be spent for improving the Traceless Genetic Programming technique and for applying it for solving other real-world problems.

\end{document}